\documentclass[10pt,twocolumn,letterpaper]{article}

\usepackage{iccv}
\usepackage{times}
\usepackage{epsfig}
\usepackage{graphicx}
\usepackage{amsmath}
\usepackage{amssymb}
\usepackage{enumitem}

\usepackage{booktabs}
\usepackage{multirow}
\usepackage{makecell}

\usepackage{color, colortbl}
\definecolor{iGray}{gray}{0.9}
\definecolor{beaublue}{rgb}{0.74, 0.83, 0.9}
\definecolor{Royal_Blue}{rgb}{0.0, 0.1, 0.66}
\usepackage[pagebackref=true,breaklinks=true,letterpaper=true,colorlinks,bookmarks=false, citecolor = Royal_Blue]{hyperref}
\usepackage{pifont}
\newcommand{\xmark}{\ding{55}}%

\usepackage{bbding}

\usepackage{graphicx}
\usepackage{mwe}
\usepackage{amsmath, bm}
\usepackage{float}
\usepackage{caption}

\newcommand{\tabincell}[2]{\begin{tabular}{@{}#1@{}}#2\end{tabular}}

\usepackage[pagebackref=true,breaklinks=true,letterpaper=true,colorlinks,bookmarks=false]{hyperref}

\iccvfinalcopy 

\def\iccvPaperID{2346} 

\ificcvfinal\fi

\begin{document}

\title{Learning Spatio-Temporal Transformer for Visual Tracking}

\author{Bin Yan$^{1,*}$, Houwen Peng$^{2, \dagger}$, Jianlong Fu$^2$, Dong Wang$^1$, Huchuan Lu$^1$ \\ $^1$Dalian University of Technology \quad $^2$Microsoft Research Asia 
}

\maketitle
\ificcvfinal\thispagestyle{empty}\fi

\begin{abstract}
In this paper, we present a new tracking architecture with an encoder-decoder transformer as the key component. The encoder models the global spatio-temporal feature dependencies between target objects and search regions, while the decoder learns a query embedding to predict the spatial positions of the target objects. 
Our method casts object tracking as a direct bounding box prediction problem, without using any proposals or predefined anchors. With the encoder-decoder transformer, the prediction of objects just uses a simple fully-convolutional network, which estimates the corners of objects directly. The whole method is end-to-end, does not need any postprocessing steps such as cosine window and bounding box smoothing, thus largely simplifying existing tracking pipelines. The proposed tracker achieves state-of-the-art performance on five challenging short-term and long-term benchmarks, while running at real-time speed, being $6\times$ faster than Siam R-CNN~\cite{SiamRCNN}. Code and models are open-sourced at \href{https://github.com/researchmm/Stark}{here}.

\end{abstract}

\newcommand\blfootnote[1]{%
\begingroup 
\renewcommand\thefootnote{}\footnote{#1}%
\addtocounter{footnote}{-1}%
\endgroup 
}
{
	
	\blfootnote{
	 $^*$Work performed when Bin is an intern of MSRA. ~$^\dagger$ Corresponding author: \href{mailto:houwen.peng@microsoft.com}{\color{black}{houwen.peng@microsoft.com}}.
	}
}


\vspace{-3mm}
\section{Introduction}



Visual object tracking is a fundamental yet challenging research topic in computer vision. Over the past few years, based on convolutional neural networks, object tracking has achieved remarkable progress \cite{SiamRPNplusplus, ATOM, SiamRCNN}. However, convolution kernels are not good at modeling long-range dependencies of image contents and features, because they only process a local neighborhood, either in space or time. Current prevailing trackers, including both the offline Siamese trackers and the online learning models, are almost all build upon convolutional operations \cite{SiameseFC,MDNet,DiMP,SiamRCNN}. 
As a consequence, these methods only perform well on modeling local relationships of image content, but being limited to capturing long-range global interactions. 
Such deficiency may degrade the model capacities on dealing with the scenarios where the global contextual information is important for localizing target objects, such as the objects undergoing large-scale variations or getting in and out of views frequently.

The problem of long range interactions has been tackled in sequence modeling through the use of transformer~\cite{transformer}. Transformer has enjoyed rich success in tasks such as natural language modeling~\cite{BERT, LanguageUML} and speech recognition~\cite{RWTHSpeech}. Recently, transformer has been employed in discriminative computer vision models and drawn great attention~\cite{ViT,DETR,Trackformer}. Inspired by the recent DEtection TRansformer (DETR) \cite{DETR},  we propose a new end-to-end tracking architecture with encoder-decoder transformer to boost the performance of conventional convolution models. 




\begin{figure}[!t]
  \begin{center}
\includegraphics[width=0.95\linewidth]{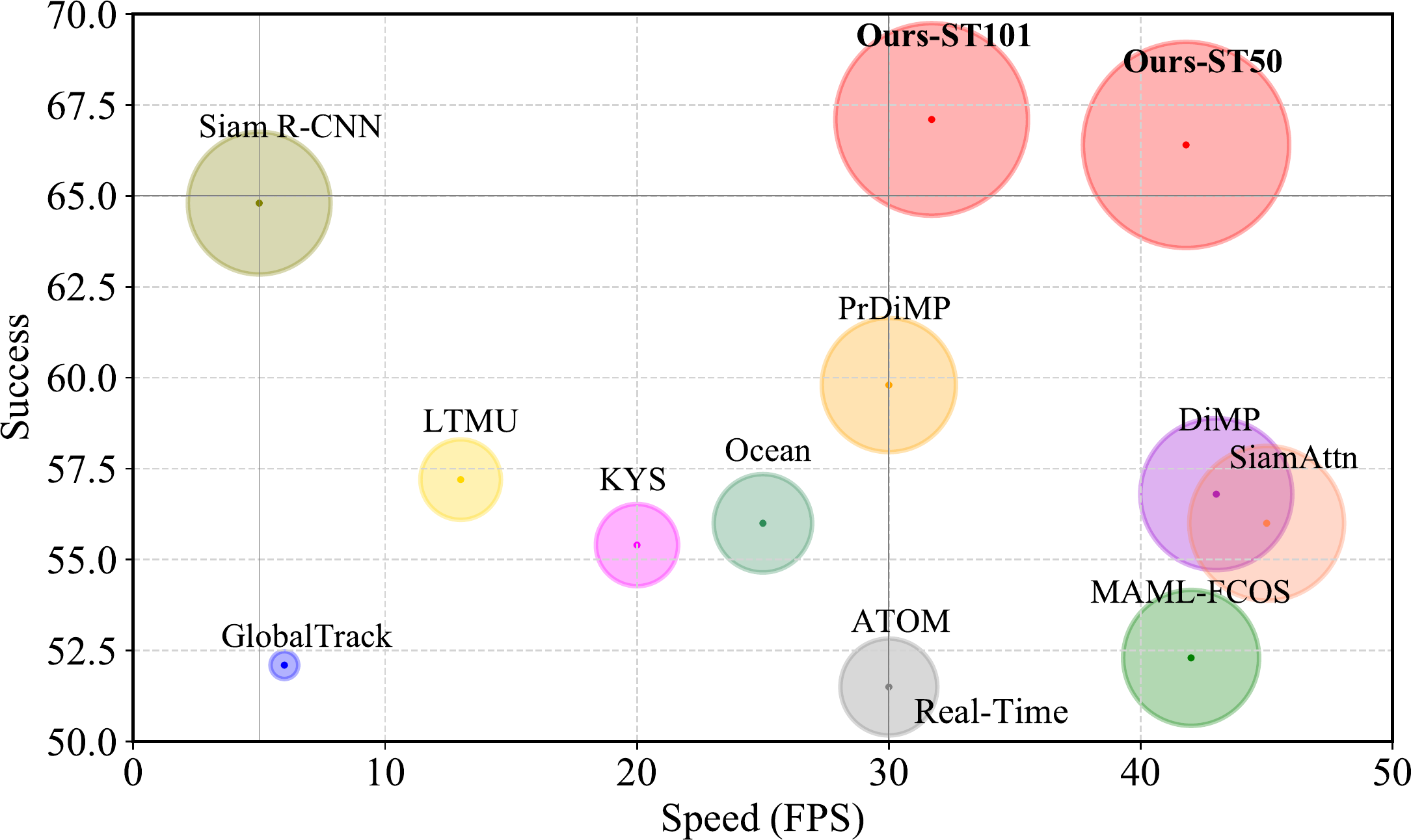}
  \end{center}
  \vspace{-5mm}
  \captionsetup{font={small}}
  \caption{Comparison with state-of-the-arts on LaSOT \cite{LaSOT}. We visualize the Success performance with respect to the Frames-Per-Seconds (\emph{fps}) tracking speed. Ours-ST101 and Ours-ST50 indicate the proposed trackers with ResNet-101 and ResNet-50 as backbones, respectively. Better viewed in color.} 
  \label{fig-page1}
\vspace{-3mm}
\end{figure}

Both spatial and temporal information are important for object tracking. The former one contains object appearance information for target localization, while the latter one includes the state changes of objects across frames.
Previous Siamese trackers~\cite{SiamRPNplusplus,SiamFC++,SiamCAR,SiamBAN} only exploit the spatial information for tracking, while online methods \cite{MemTrack,UpdateNet,ATOM,DiMP} use historical  predictions for model updates. Although being successful, these methods do not explicitly model the relationship between space and time. In this work, considering the superior capacity on modeling global dependencies, we resort to transformer to integrate spatial and temporal information for tracking, generating discriminative spatio-temporal features for object localization.

More specifically, we propose a new spatio-temporal architecture based on the encoder-decoder transformer for visual tracking. The new architecture contains three key components: an encoder, a decoder and a prediction head. The encoder accepts inputs of an initial target object, the current image, and a dynamically updated template. The self-attention modules in the encoder learn the relationship between the inputs through their feature dependencies. Since the template images are updated throughout video sequences, the encoder can capture both spatial and temporal information of the target. The decoder learns a query embedding to predict the spatial positions of the target object. A corner-based prediction head is used to estimate the bounding box of the target object in the current frame. Meanwhile, a score head is learned to control the updates of the dynamic template images.



Extensive experiments demonstrate that our method establishes new state-of-the-art performance on on both short-term~\cite{GOT10K, trackingnet} and long-term tracking benchmarks~\cite{LaSOT, vot2020}. For instance, our spatio-temporal transformer tracker surpasses Siam R-CNN \cite{SiamRCNN} by 3.9\% (AO score) and 2.3\% (Success) on GOT-10K~\cite{GOT10K} and LaSOT \cite{LaSOT}, respectively. It is also worth noting that compared with previous long-term trackers~\cite{LTMU,SiamRCNN,SPLT}, the framework of our method is much simpler. Specifically, previous methods usually consist of multiple components, such as base trackers~\cite{ATOM, SiamMask}, target verification modules~\cite{RTMDNet}, and global detectors~\cite{FasterRCNN, GlobalTrack}. In contrast, our method only has a single network 
learned in an end-to-end fashion. Moreover, 
our tracker can run at real-time speed, being $6\times$ faster than Siam R-CNN (30 v.s. 5 \emph{fps}) on a Tesla V100 GPU, as shown in Fig.~\ref{fig-page1}


In summary, this work has three contributions.
\begin{itemize}
	\vspace{-1mm}
	\item{We propose a new transformer architecture dedicated to visual tracking. It is capable of capturing global feature dependencies of both spatial and temporal information in video sequences.}
	\vspace{-1mm}
	\item{
	The whole method is end-to-end, does not need any postprocessing steps such as cosine window and bounding box smoothing, thus largely simplifying existing tracking pipelines.
	}
	\vspace{-1mm}
	\item{The proposed trackers achieve state-of-the-art performance on five challenging short-term and long-term benchmarks, while running at real-time speed.}
\end{itemize}

\section{Related Work}
\textbf{Transformer in Language and Vision.} Transformer is originally proposed by Vaswani \emph{et al.}~\cite{transformer} for machine translation task, and has became a prevailing architecture in language modeling. Transformer takes a sequence as the input, scans through each element in the sequence and learns their dependencies. This feature makes transformer be intrinsically good at capturing global information in sequential data. Recently, transformer has shown their great potential in vision tasks like image classification~\cite{ViT}, object detection~\cite{DETR}, semantic segmentation~\cite{Max-Deeplab}, multiple object tracking~\cite{TransTrack, Trackformer}, etc. 
Our work is inspired by the recent work DETR~\cite{DETR}, but has following fundamental differences. (1) The studied tasks are different. DETR is designed for object detection, while this work is for object tracking. (2) The network inputs are different. DETR takes the whole image as the input, while our input is a triplet consisting of one search region and two templates. Their features from the backbone are first flatten and concatenated then sent to the encoder. (3) The query design and training strategies are different. DETR uses 100 object queries and uses the Hungarian algorithm to match predictions with ground-truths during training. In contrast, our method only uses one query and always matches it with the ground-truth without using the Hungarian algorithm. (4) The bounding box heads are different. DETR uses a three-layer perceptron to predict boxes. Our network adopts a corner-based box head for higher-quality localization.

Moreover, TransTrack~\cite{TransTrack} and TrackFormer~\cite{Trackformer} are two most recently representative works on transformer tracking. TransTrack~\cite{TransTrack} has the following features. (1) The encoder takes the image features of both the current and the previous frame as the inputs. (2)
It has two decoders, which take the learned object queries and queries from the last frame as the input respectively. With different queries, the output sequence from the encoder are transformed into detection boxes and tracking boxes respectively. (3) The predicted two groups of boxes are matched based on the IoUs using the Hungarian algorithm~\cite{hungarian}.
While Trackformer~\cite{Trackformer} has the following features. (1) It only takes the current frame features as the encoder inputs. (2) There is only one decoder, where the learned object queries and the track queries from the last frame interact with each other. (3) 
It associates tracks over time solely by attention operations, not relying on any additional matching such as motion or appearance modeling. 
In contrast, our work has the following fundamental differences with these two methods. (1) Network inputs are different. Our input is a triplet consisting of the current search region, the initial template and a dynamic template. (2) Our method captures the appearance changes of the tracked targets by updating the dynamic template, rather than updating object queries as~\cite{TransTrack, Trackformer}.

\textbf{Spatio-Temporal Information Exploitation.} 
Exploitation of spatial and temporal information is a core problem in object tracking field. Existing trackers can be divided into two classes: spatial-only ones and spatio-temporal ones. Most of offline Siamese trackers~\cite{SiameseFC, SiameseRPN, SiamRPNplusplus, Ocean, PG-Net} belong to the spatial-only ones, which consider the object tracking as a template-matching between the initial template and the current search region. To extract the relationship between the template and the search region along the spatial dimension, most trackers adopt the variants of correlation, including the naive correlation~\cite{SiameseFC, SiameseRPN}, the depth-wise correlation~\cite{SiamRPNplusplus, Ocean}, and the point-wise correlation~\cite{PG-Net, AlphaRefine}. Although achieving remarkable progress in recent years, these methods merely capture local similarity, while ignoring global information. 
By contrast, the self-attention mechanism in transformer can capture long-range relationship, making it suitable for pair-wise matching tasks. Compared with spatial-only trackers, spatio-temporal ones additionally exploit temporal information to improve trackers' robustness. These methods can also be divided into two classes: gradient-based and gradient-free ones. Gradient-based methods require gradient computation during inference. One of the classical works is MDNet~\cite{MDNet}, which updates domain-specific layers with gradient descent. To improve the optimization efficiency, later works~\cite{ATOM, DiMP, GradNet, MAML-track, ROAM} adopt more advanced optimization methods like Gauss-Newton method or meta-learning-based update strategies. 
However, many real-world devices for deploying deep learning do not support back-propagation, which restricts the application of gradient-based methods. In contrast, gradient-free methods have larger potentials in real-world applications. One class of gradient-free methods~\cite{MemTrack, UpdateNet} exploits an extra network to update the template of Siamese trackers~\cite{SiameseFC, DSiam}. Another representative work LTMU~\cite{LTMU} learns a meta-updater to predict whether the current state is reliable enough to be used for the update in long-term tracking. Although being effective, these methods cause the separation between space and time. In contrast, our method integrates the spatial and temporal information as a whole, simultaneously learning them with the transformer.

\textbf{Tracking Pipeline and Post-processing.} The tracking pipelines of previous trackers~\cite{SiamRPNplusplus, SiamFC++, Ocean, SiamRCNN} are complicated. Specifically, they first generate a large number of box proposals with confidence scores, then use various post-processing to choose the best bounding box as the tracking result. The commonly used post-processing includes cosine window, scale or aspect-ratio penalty, bounding box smoothing, tracklet-based dynamic programming, etc. Though it brings better results, post-processing causes the performance being sensitive to hyper-parameters. There are some trackers~\cite{GOTURN, GlobalTrack} attempting to simplify the tracking pipeline, but their performances still lag far behind that of state-of-the-art trackers. This work attempts to close this gap, achieving top performance by predicting one single bounding box in each frame.

\vspace{-2mm}
\section{Method}

\vspace{-2mm}
In this section, we propose the \textbf{s}patio-\textbf{t}emporal tr\textbf{a}ns-fo\textbf{r}mer network for visual trac\textbf{k}ing, called STARK. 
For clarity, we first introduce a simple baseline method that directly applies the original encoder-decoder transformer for tracking. The baseline method only considers spatial information and achieves impressive performance. After that, we extend the baseline to learn both spatial and temporal representations for target localization. We introduce an  dynamic template and an update controller to capture the appearance changes of target objects. 

\begin{figure}[!t]
  \begin{center}
  \includegraphics[width=0.9\linewidth]{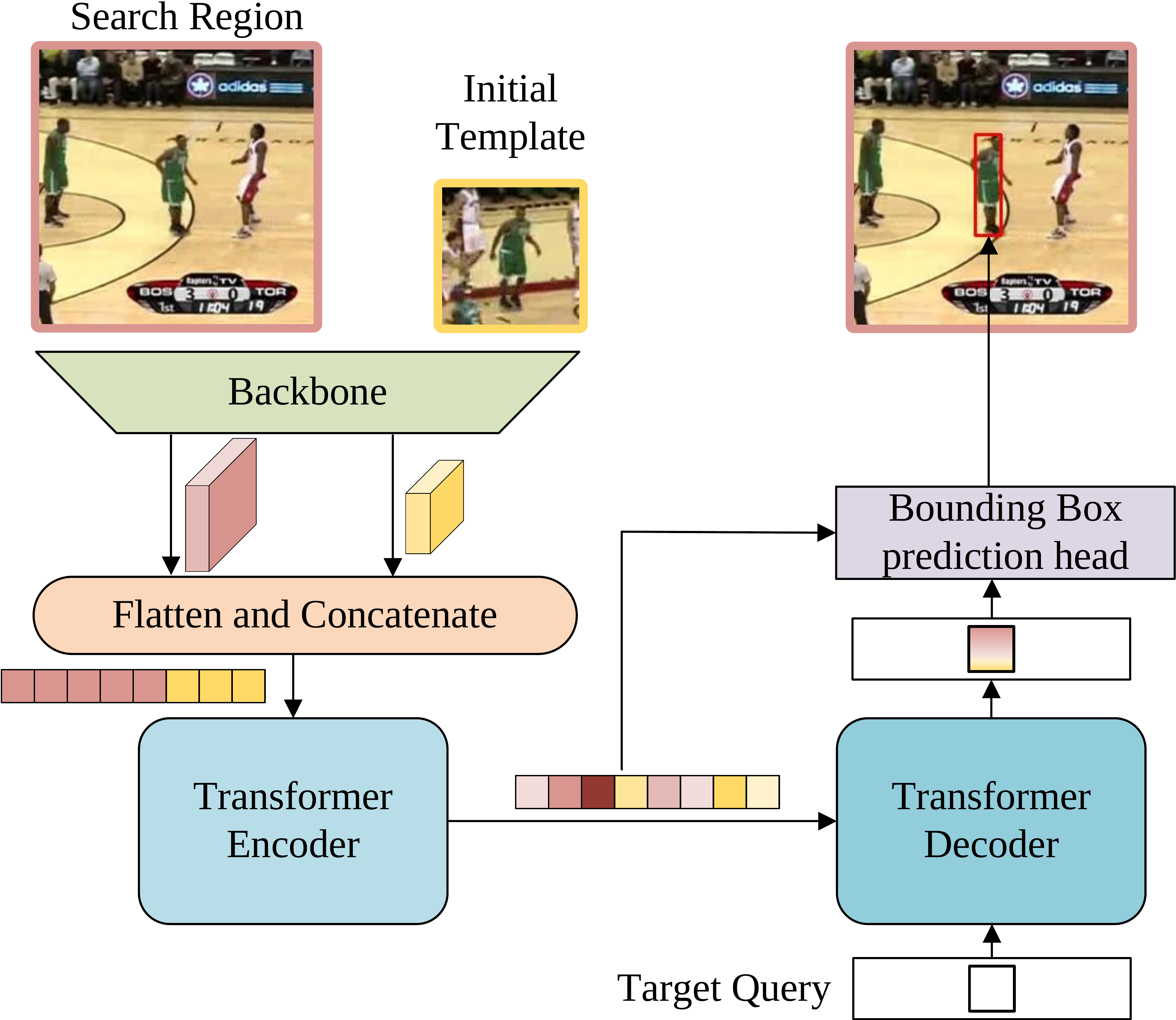}
  \end{center}
  \vspace{-5mm}
  \caption{Framework for spatial-only tracking.}
  \label{fig-framework}
\vspace{-5mm}
\end{figure}

\subsection{A Simple Baseline Based on Transformer} \label{sec-spatial}

We present a simple baseline framework based on visual transformer for object tracking. The network architecture is demonstrated in Fig.~\ref{fig-framework}. It mainly consists of three components: a convolutional backbone, an encoder-decoder transformer, and a bounding box prediction head. 


\textbf{Backbone}. Our method can use arbitrary convolutional networks as the backbone for feature extraction. Without loss of generality, we adopt the vanilla  ResNet~\cite{ResNet} as the backbone. More concretely, except for removing the last stage and fully-connected layers, there is no other change for the original ResNet~\cite{ResNet}. The input of the backbone is a pair of images: a template image of the initial target object $z \in \mathbb{R}^{3\times H_z \times W_z}$ and a search region of the current frame $x \in \mathbb{R}^{3\times H_x \times W_x}$. After passing through of the backbone, the template $z$ and the search image $x$ are mapped to two feature maps $f_z \in \mathbb{R}^{C\times \frac{H_z}{s} \times \frac{W_z}{s}}$ and $f_x \in \mathbb{R}^{C\times \frac{H_x}{s} \times \frac{W_x}{s}}$. 


\textbf{Encoder}. The feature maps output from the backbone require pre-processing before feeding into the encoder. To be specific, a bottleneck layer is first used to reduce the channel number from $C$ to $d$. Then the feature maps are flatten and concatenated along the spatial dimension, producing a feature sequence with length of $ \frac{H_z}{s}\frac{W_z}{s}+\frac{H_x}{s}\frac{W_x}{s}$ and dimension of $d$, which servers as the input for the transformer encoder. The encoder consists of $N$ encoder layers, each of which is made up of a multi-head self-attention module with a feed-forward network. Due to the permutation-invariance of the original transformer \cite{transformer}, we add sinusoidal positional embeddings to the input sequence. The encoder captures the feature dependencies among all elements in the sequence and reinforces the original features with global contextual information, thus allowing the model to learn discriminative features for object localization. 

\textbf{Decoder}. The decoder takes a target query and the enhanced feature sequence from the encoder as the input. Different from DETR~\cite{DETR} adopting 100 object queries, we only input one single query into the decoder to predict one bounding box of the target object. Besides, since there is only one prediction, we remove the Hungarian algorithm \cite{hungarian} used in DETR for prediction association. Similar to the encoder, the decoder stacks $M$ decoder layers, each of which consists of a self-attention, an encoder-decoder attention, and a feed-forward network. In the encoder-decoder attention module, the target query can attend to all positions on the template and the search region features, thus learning robust representations for the final bounding box prediction.

\begin{figure}[!t]
  \begin{center}
  \includegraphics[width=1.0\linewidth]{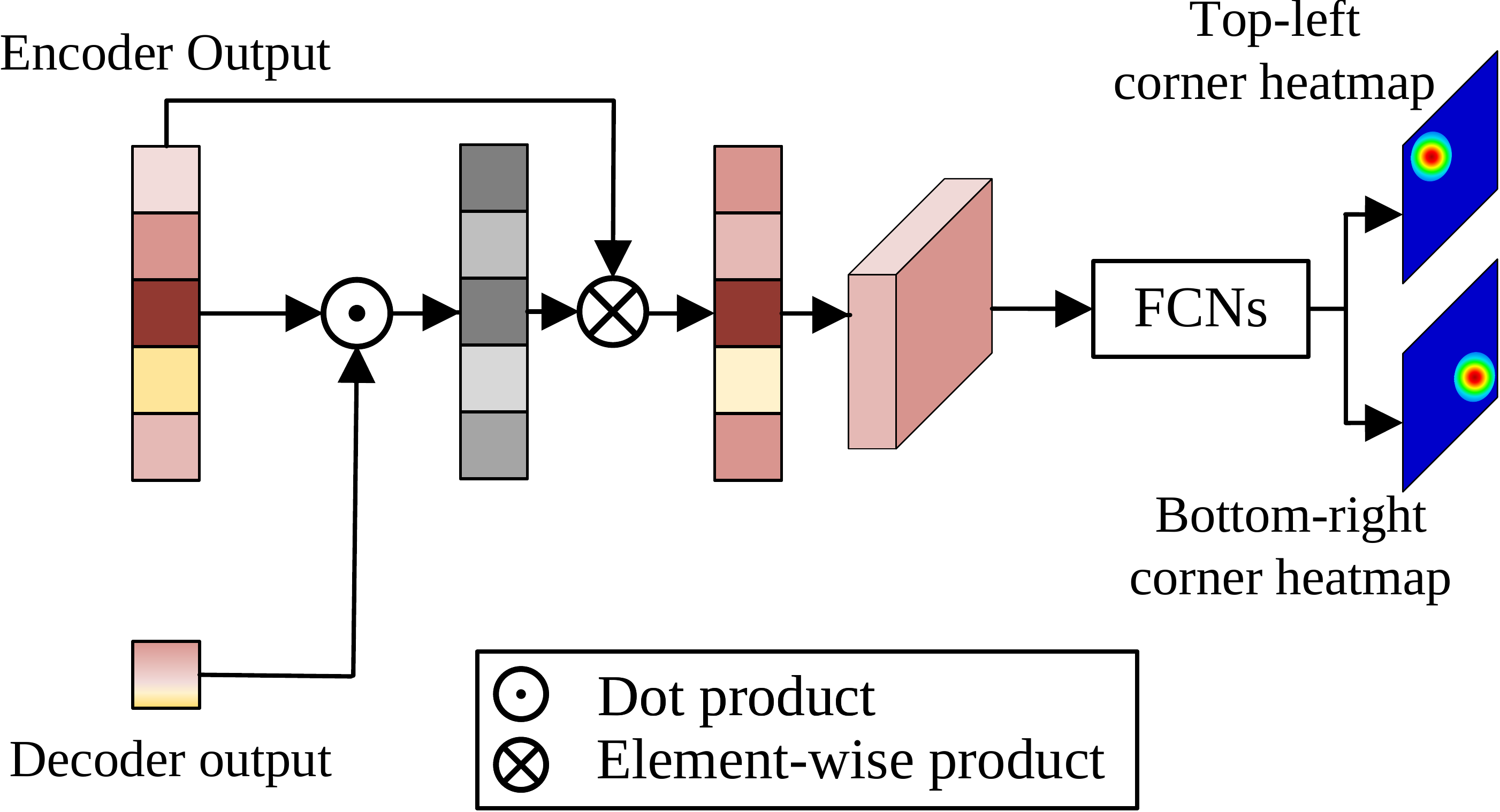}
  \end{center}
  \vspace{-5mm}
  \caption{Architecture of the box prediction head.}
  \label{fig-corner-head}
\vspace{-3mm}
\end{figure}

\textbf{Head}. DETR~\cite{DETR} adopts a three-layer perceptron to predict object box coordinates. However, as pointed by GFLoss~\cite{GfocalLoss}, directly regressing the coordinates is equivalent to fitting a Dirac delta distribution, which fails to consider the ambiguity and uncertainty in the datasets. This representation is not flexible and not robust to challenges such as occlusion and cluttered background in object tracking. To improve the box estimation quality, we design a new  prediction head through estimating the probability distribution of the box corners. As shown in Fig.~\ref{fig-corner-head}, we first take the search region features from the encoder's output sequence, then compute the similarity between the search region features and the output embedding from the decoder. Next, the similarity scores are element-wisely multiplied with the search region features to enhance important regions and weaken the less discriminative ones. The new feature sequence is reshaped to a feature map $f \in \mathbb{R}^{d\times \frac{H_s}{s} \times \frac{W_s}{s}}$, and then fed into a simple fully-convolutional network (FCN). The FCN consists of $L$ stacked Conv-BN-ReLU layers and outputs two probability maps $P_{tl}(x,y)$ and $P_{br}(x,y)$ for the top-left and the bottom-right corners of object bounding boxes, respectively. Finally, the predicted box coordinates $(\widehat{x_{tl}},\ \widehat{y_{tl}})$ and $(\widehat{x_{br}},\ \widehat{y_{br}})$ are obtained by computing the expectation of corners' probability distribution as shown in Eq.~(\ref{eqn1}). Compared with DETR, our method explicitly models uncertainty in the coordinate estimation, generating more accurate and robust predictions for object tracking.
\begin{equation}
	\label{eqn1}
	\begin{aligned}
		(\widehat{x_{tl}},\ \widehat{y_{tl}})=(\sum_{y=0}^{H}\sum_{x=0}^{W}{{x}\cdot{P_{tl}(x,y)}},\ \sum_{y=0}^{H}\sum_{x=0}^{W}{{y}\cdot{P_{tl}(x,y)}}),\\
		(\widehat{x_{br}},\ \widehat{y_{br}})=(\sum_{y=0}^{H}\sum_{x=0}^{W}{{x}\cdot{P_{br}(x,y)}},\ \sum_{y=0}^{H}\sum_{x=0}^{W}{{y}\cdot{P_{br}(x,y)}}),
	\end{aligned}
\end{equation}

\textbf{Training and Inference}. Our baseline tracker is trained in an end-to-end fashion with the combination of the $\ell_1$ Loss and the generalized IoU loss~\cite{GIoULoss} as in DETR. The loss function can be written as 
\begin{equation}
	\label{equ-loss-loc}
	\begin{aligned}
		L=\lambda_{iou}L_{iou}(b_i,{\hat{b}}_i)+\lambda_{L_1}L_1(b_i,\hat{b}_i).
	\end{aligned}
\end{equation}
where $b_i$ and $\hat{b}_i$ represent the groundtruth and the predicted box respectively and $\lambda_{iou},\lambda_{L_1}\in \mathbb{R}$ are hyperparameters. 
But unlike DETR, we do not use the classification loss and the Hungarian algorithm, thus further simplifying the training process. During inference, the template image together with its features from the backbone are initialized by the first frame and fixed in the subsequent frames. During tracking, in each frame, the network takes a search region from the current frame as the input, and returns the predicted box as the final result, without using any post-processing such as cosine window or bounding box smoothing.

\subsection{Spatio-Temporal Transformer Tracking}
\label{sec-temporal}

\begin{figure}[!t]
  \begin{center}
  \includegraphics[width=1.0\linewidth]{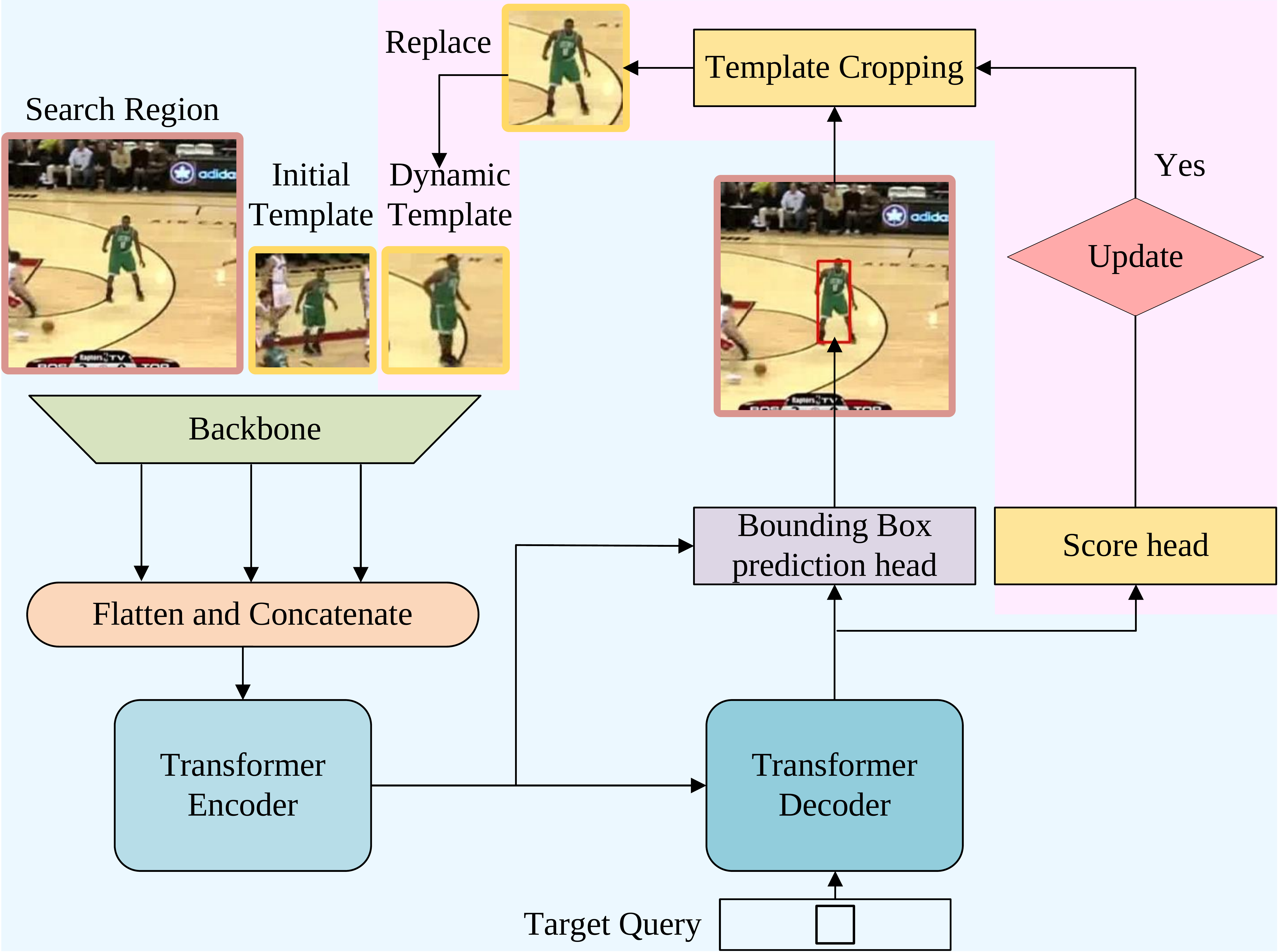}
  \end{center}
  \vspace{-5mm}
  \caption{Framework for spatio-temporal tracking. The differences with the spatial-only architecture are highlighted in pink.}
  \label{fig-framework-temporal}
\vspace{-4mm}
\end{figure}

Since the appearance of a target object may change significantly as time proceeds, it is important to capture the latest state of the target for tracking. In this section, we demonstrate how to exploit spatial and temporal information simultaneously based on the previously introduced baseline. Three key differences are made, including the network inputs, an extra score head, and the training \& inference strategy. We elaborate them one by one as below. The spatio-temporal architecture is shown in Fig.~\ref{fig-framework-temporal}.

\textbf{Input}. Different from the baseline method which only uses the first and the current frames, the spatio-temporal method introduces an dynamically updated template sampled from intermediate frames as an additional input, as shown in Fig.~\ref{fig-framework-temporal}. 
Beyond the spatial information from the initial template, the dynamic template can captures the target appearance changes with time, providing additional temporal information. Similar to the baseline architecture in Sec.~\ref{sec-spatial}, feature maps of the triplet are flatten and concatenated then sent to the encoder. The encoder extracts discriminative spatio-temporal features by modeling the global relationships among all elements in both spatial and temporal dimensions.



\textbf{Head}. During tracking, there are some cases where the dynamic template should not be updated. For example, the cropped template is not reliable when the target has been completely occluded or has moved out of view, or when the tracker has drifted. For simplicity, we consider that the dynamic template could be updated as long as the search region contains the target. To automatically determine whether the current state is reliable, we add a simple score prediction head, which is a three-layer perceptron followed by a sigmoid activation. The current state is considered reliable if the score is higher than 
the threshold $\tau$.

\textbf{Training and Inference}. As pointed out by recent works~\cite{RevisitingRCNN,TSD}, jointly learning of localization and classification may cause sub-optimal solutions for both tasks, and it is helpful to decouple localization and classification. 
Therefore, we divide the training process into two stages, regarding the localization as a primary task and the classification as a secondary task. To be specific, in the first stage, the whole network, except for the score head, is trained end-to-end only with the localization-related losses in Eq.~\ref{equ-loss-loc}. In this stage, we ensure all search images to contain the target objects and let the model learn the localization capacity. In the second stage, only the score head is optimized with binary cross-entropy loss defined as 
\begin{equation}
	\label{equ-loss-ce}
	\begin{aligned}
L_{ce}=y_ilog\left(P_i\right)+\left(1-y_i\right)log\left(1-P_i\right),
	\end{aligned}
\end{equation}
where $y_i$ is the groundtruth label and $P_i$ is the predicted confidence
, and all other parameters are frozen to avoid affecting the localization capacity. In this way, the final model learn both localization and classification capabilities after the two-stage training.

During inference, two templates and corresponding features are initialized in the first frame. Then a search region is cropped and fed into the network, generating one bounding box and a confidence score. The dynamic template is updated only when the update interval is reached and the confidence score is higher than the threshold $\tau$. For efficiency, we set the update interval as $T_u$ frames. The new template is cropped from the original image and then fed into the backbone for feature extraction. 

\section{Experiments}

This section first presents the implementation details and the results of our STARK tracker on multiple benchmarks, with comparisons to state-of-the-art methods. Then, ablation studies are presented to analyze the effects of the key components in the proposed networks. We also report the results of other candidate frameworks and compare them with our method to demonstrate its superiority. Finally, visualization on attention maps of the encoder and the decoder are provided to understand how the transformer works.

\subsection{Implementation Details}
Our trackers are implemented using Python 3.6 and PyTorch 1.5.1. The experiments are conducted on a server with 8 16GB Tesla V100 GPUs.

\textbf{Model}. We report the results of three variants of STARK: STARK-S50, STARK-ST50 and STARK-ST101. STARK-S50 only exploits spatial information and takes ResNet-50~\cite{ResNet} as the backbone, \emph{i.e.}, the baseline tracker presented in Sec. \ref{sec-spatial}. STARK-ST50 and STARK-ST101 take ResNet-50 and ResNet-101 as the backbones respectively, exploiting both spatial and temporal information, \emph{i.e.}, the spatio-temporal tracker presented in Sec. \ref{sec-temporal}. 

The backbones are initialized with the parameters pretrained on ImageNet. The BatchNorm \cite{BatchNormalization} layers are frozen during training. Backbone features are pooled from the fourth stage with a stride of 16. The transformer architecture is similar to that in DETR~\cite{DETR} with 6 encoder layers and 6 decoder layers, which consist of multi-head attention layers (MHA) and feed-forward networks (FFN). The MHA have 8 heads, width 256, while the FFN have hidden units of 2048. Dropout ratio of 0.1 is used. The bounding box prediction head is a lightweight FCN, consisting of 5 stacked Conv-BN-ReLU layers. The classification head is a three-layer perceptron with 256 hidden units in each layer.

\textbf{Training}. The training data consists of the train-splits of LaSOT~\cite{LaSOT}, GOT-10K~\cite{GOT10K}, COCO2017~\cite{COCO}, and TrackingNet~\cite{trackingnet}. As required by VOT2019 challenge, we remove 1k forbidden sequences from GOT-10K training set. The sizes of search images and templates are $320\times320$ pixels and $128\times128$ pixels respectively, corresponding to $5^2$ and $2^2$ times of the target box area. Data augmentations, including horizontal flip and brightness jittering, are used. The minimal training data unit for STARK-ST is one triplet, consisting of two templates and one search images. The whole training process of STARK-ST consists of two stages, which take 500 epochs for localization and 50 epochs for classification, respectively. Each epoch uses $6\times10^4$ triplets. The network is optimized using AdamW optimizer~\cite{AdamW} and weight decay $10^{-4}$. The loss weights $\lambda_{L1}$ and $\lambda_{iou}$ are set to 5 and 2 respectively. Each GPU hosts 16 triplets, hence the mini-batch size is 128 triplets per iteration. The initial learning rates of the backbone and the rest parts are $10^{-5}$ and $10^{-4}$ respectively. The learning rate drops by a factor of 10 after 400 epochs in the first stage and after 40 epochs in the second stage. The training setting of STARK-S is almost the same as that of STARK-ST, except that (1) the minimal training data unit of STARK-S is a template-search pair; (2) the training process only has the first stage.

\textbf{Inference}. The dynamic template update interval $T_u$ and the confidence threshold $\tau$ are respectively set to 200 frames and 0.5 by default. The inference pipeline only contains a forward pass and a coordinate transformation from the search region to the original image, without any extra post-processing.

\begin{table*}[th]
\small
    \centering
    \caption{
    \small
    Comparisons on GOT-10k test set~\cite{GOT10K}}
    \vspace{-2mm}
    \resizebox{\linewidth}{!}{
    \begin{tabular}{c|cccccccccccc}
        \hline
  \small
        &\tabincell{c}{SiamFC\\~\cite{SiameseFC}}&\tabincell{c}{SiamFCv2\\~\cite{CFNet}}&\tabincell{c}{ATOM\\~\cite{ATOM}}&\tabincell{c}{SiamFC++\\~\cite{SiamFC++}}&\tabincell{c}{D3S\\~\cite{D3S}}&\tabincell{c}{DiMP50\\~\cite{DiMP}}&\tabincell{c}{Ocean\\~\cite{Ocean}}&\tabincell{c}{PrDiMP50\\~\cite{PrDiMP}}&\tabincell{c}{SiamRCNN\\~\cite{SiamRCNN}}&\tabincell{c}{\textbf{STARK}\\\textbf{-S50}}&\tabincell{c}{\textbf{STARK}\\\textbf{-ST50}}&\tabincell{c}{\textbf{STARK}\\\textbf{-ST101}}\\
        \hline
        AO(\%)&34.8&37.4&55.6&59.5&59.7&61.1&61.1&63.4&64.9&\textbf{\textcolor[rgb]{0,0,1}{67.2}}&\textbf{\textcolor[rgb]{0,1,0}{68.0}}&\textbf{\textcolor[rgb]{0.8,0.2,0}{68.8}} \\
        SR0.5(\%)&35.3&40.4&63.4&69.5&67.6&71.7&72.1&73.8&72.8&\textbf{\textcolor[rgb]{0,0,1}{76.1}}&\textbf{\textcolor[rgb]{0,1,0}{77.7}}&\textbf{\textcolor[rgb]{0.8,0.2,0}{78.1}} \\
        SR0.75(\%)&9.8&14.4&40.2&47.9&46.2&49.2&47.3&54.3&59.7&\textbf{\textcolor[rgb]{0,0,1}{61.2}}&\textbf{\textcolor[rgb]{0,1,0}{62.3}}&\textbf{\textcolor[rgb]{0.8,0.2,0}{64.1}} \\
        \hline
    \end{tabular}
    }
    \label{tab-got10k}
\vspace{-3mm}
\end{table*}

\begin{table*}[!th]
\small
    \centering
    \caption{\small Comparisons on TrackingNet test set~\cite{trackingnet}.}
    \vspace{-2mm}
    \resizebox{\linewidth}{!}{
    \begin{tabular}{c|cccccccccccc}
        \hline
\small
        &\tabincell{c}{DSiamRPN\\~\cite{DSiam}}&\tabincell{c}{ATOM\\~\cite{ATOM}}&\tabincell{c}{SiamRPN++\\~\cite{SiamRPNplusplus}}&\tabincell{c}{DiMP50\\~\cite{DiMP}}&\tabincell{c}{SiamAttn\\~\cite{Deform_siam}}&\tabincell{c}{SiamFC++\\~\cite{SiamFC++}}&\tabincell{c}{MAML-FCOS\\~\cite{MAML-track}}&\tabincell{c}{PrDiMP50\\~\cite{PrDiMP}}&\tabincell{c}{SiamRCNN\\~\cite{SiamRCNN}}&\tabincell{c}{\textbf{STARK}\\\textbf{-S50}}&\tabincell{c}{\textbf{STARK}\\\textbf{-ST50}}&\tabincell{c}{\textbf{STARK}\\\textbf{-ST101}}\\
        \hline
        AUC(\%)&63.8&70.3&73.3&74.0&75.2&75.4&75.7&75.8&\textbf{\textcolor[rgb]{0,0,1}{81.2}}&80.3&\textbf{\textcolor[rgb]{0,1,0}{81.3}}&\textbf{\textcolor[rgb]{0.8,0.2,0}{82.0}}\\
        $P_{norm}$(\%)&73.3&77.1&80.0&80.1&81.7&80.0&82.2&81.6&\textbf{\textcolor[rgb]{0,0,1}{85.4}}&85.1&\textbf{\textcolor[rgb]{0,1,0}{86.1}}&\textbf{\textcolor[rgb]{0.8,0.2,0}{86.9}}\\
        \hline
    \end{tabular}
    }
    \label{tab-trackingnet}
\vspace{-3mm}
\end{table*}

\begin{table*}[!th]
\small
    \centering
    \caption{\small
 Comparisons on VOT2020~\cite{vot2020}.``+AR" means using Alpha-Refine to predict masks. The upper row summarizes trackers that only predict bounding boxes and the lower row presents trackers that report masks.}
    \vspace{-3mm}
    \resizebox{\linewidth}{!}{
    \begin{tabular}{c|cccccccccccc}
        \hline
        &\tabincell{c}{IVT\\~\cite{IVT}}&\tabincell{c}{KCF\\~\cite{KCF}}&\tabincell{c}{SiamFC\\~\cite{SiameseFC}}&\tabincell{c}{CSR-DCF\\~\cite{CSR-DCF}}&\tabincell{c}{ATOM\\~\cite{ATOM}}&\tabincell{c}{DiMP\\~\cite{DiMP}}&\tabincell{c}{UPDT\\~\cite{UPDT}}&\tabincell{c}{DPMT\\\quad}&\tabincell{c}{SuperDiMP\\~\cite{SuperDiMP}}&\tabincell{c}{\textbf{STARK}\\\textbf{-S50}}&\tabincell{c}{\textbf{STARK}\\\textbf{-ST50}}&\tabincell{c}{\textbf{STARK}\\\textbf{-ST101}}\\
        \hline
        EAO($\uparrow$)&0.092&0.154&0.179&0.193&0.271&0.274&0.278&0.303&0.305&0.280&0.308&0.303 \\
        Accuracy($\uparrow$)&0.345&0.407&0.418&0.406&0.462&0.457&0.465&0.492&0.477&0.477&0.478&0.481 \\
        Robustness($\uparrow$)&0.244&0.432&0.502&0.582&0.734&0.740&0.755&0.745&0.786&0.728&0.799&0.775 \\
        \hline
        &\tabincell{c}{STM\\~\cite{STM}}&\tabincell{c}{SiamEM\\\quad}&\tabincell{c}{SiamMask\\~\cite{SiamMask}}&\tabincell{c}{SiamMargin\\~\cite{vot2020}}&\tabincell{c}{Ocean\\~\cite{Ocean}}&\tabincell{c}{D3S\\~\cite{D3S}}&\tabincell{c}{FastOcean\\\quad}&\tabincell{c}{AlphaRef\\~\cite{vot2020}}&\tabincell{c}{OceanPlus\\~\cite{OceanPlus}}&\tabincell{c}{\textbf{STARK}\\\textbf{-S50+AR}}&\tabincell{c}{\textbf{STARK}\\\textbf{-ST50+AR}}&\tabincell{c}{\textbf{STARK}\\\textbf{-ST101+AR}}\\
        \hline
        EAO($\uparrow$)&0.308&0.310&0.321&0.356&0.430&0.439&0.461&0.482&\textbf{\textcolor[rgb]{0,0,1}{0.491}}&0.462&\textbf{\textcolor[rgb]{0.8,0.2,0}{0.505}}&\textbf{\textcolor[rgb]{0,1,0}{0.497}} \\
        Accuracy($\uparrow$)&0.751&0.520&0.624&0.698&0.693&0.699&0.693&0.754&0.685&\textbf{\textcolor[rgb]{0,1,0}{0.761}}&\textbf{\textcolor[rgb]{0,0,1}{0.759}}&\textbf{\textcolor[rgb]{0.8,0.2,0}{0.763}} \\
        Robustness($\uparrow$)&0.574&0.743&0.648&0.640&0.754&0.769&\textbf{\textcolor[rgb]{0,0,1}{0.803}}&0.777&\textbf{\textcolor[rgb]{0.8,0.2,0}{0.842}}&0.749&\textbf{\textcolor[rgb]{0,1,0}{0.817}}&0.789 \\
        \hline
    \end{tabular}
    }
    \label{tab-vot}
\vspace{-3mm}
\end{table*}

\begin{table*}[th]
\small
    \centering
    \caption{
    \small
    Comparisons on VOT-LT2020 benchmark~\cite{vot2020}}
    \vspace{-2mm}
    \resizebox{\linewidth}{!}{
    \begin{tabular}{c|cccccccccc}
        \hline
  \small
        &SPLT~\cite{SPLT}&ltMDNet&SiamDW\_LT~\cite{Deeper-wider-SiamRPN}&RLT\_DiMP&CLGS&Megtrack&LTMU\_B~\cite{LTMU}&LT\_DSE&\textbf{STARK-ST50}&\textbf{STARK-ST101}\\
        \hline
        F-score(\%)&56.5&57.4&65.6&67.0&67.4&68.7&69.1&\textbf{\textcolor[rgb]{0,0,1}{69.5}}&\textbf{\textcolor[rgb]{0.8,0.2,0}{70.2}}&\textbf{\textcolor[rgb]{0,1,0}{70.1}} \\
        Pr(\%)&58.7&64.9&67.8&65.7&\textbf{\textcolor[rgb]{0.8,0.2,0}{73.9}}&70.3&70.1&\textbf{\textcolor[rgb]{0,1,0}{71.5}}&\textbf{\textcolor[rgb]{0,0,1}{71.0}}&70.2 \\
        Re(\%)&54.4&51.4&63.5&68.4&61.9&67.1&68.1&\textbf{\textcolor[rgb]{0,0,1}{67.7}}&\textbf{\textcolor[rgb]{0,1,0}{69.5}}&\textbf{\textcolor[rgb]{0.8,0.2,0}{70.1}} \\
        \hline
    \end{tabular}
    }
    \label{tab-vot20lt}
\vspace{-3mm}
\end{table*}


\begin{figure}[!t]
  \begin{center}
  \begin{tabular}{cc}
  \hspace{-5mm}\includegraphics[width=0.5\linewidth]{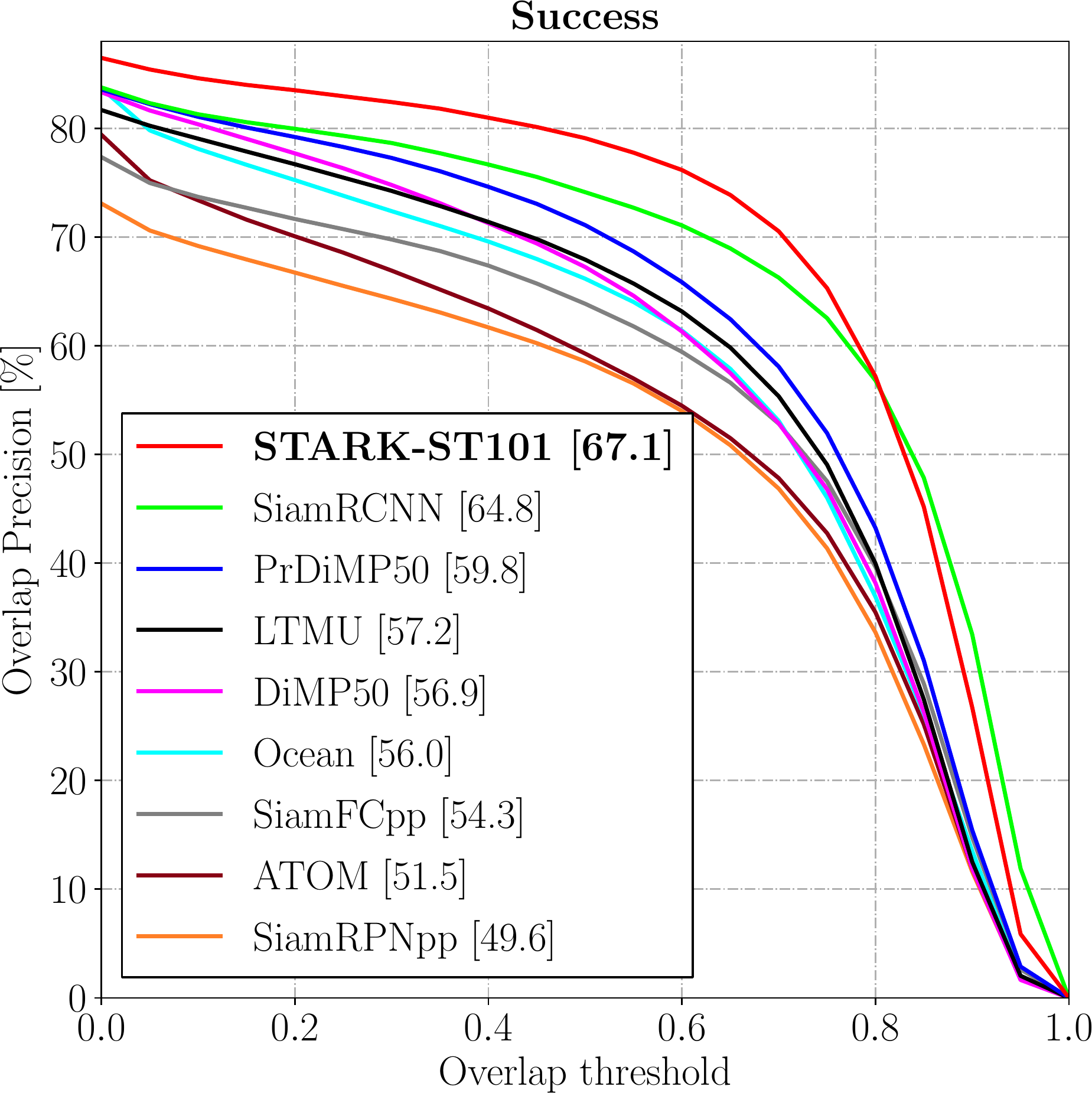} \ &\hspace{-5mm}
  \includegraphics[width=0.5\linewidth]{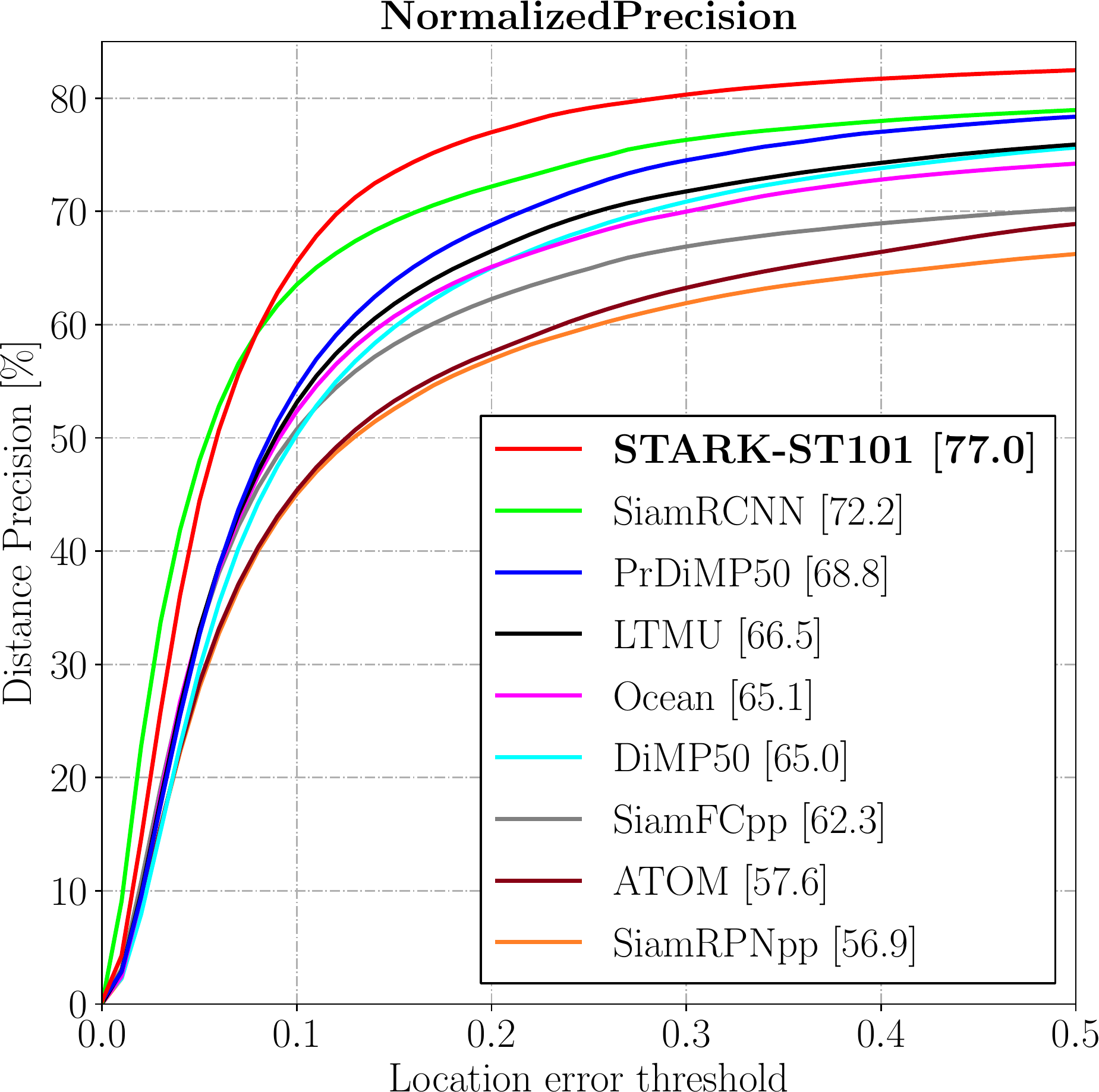}\\
  \end{tabular}
  \end{center}
  \vspace{-5mm}
  \caption{Comparisons on LaSOT test set~\cite{LaSOT}.}
  \label{fig-lasot}
\vspace{-4mm}
\end{figure}

\subsection{Results and Comparisons}\label{sec-main-results}
We compare our STARK with existing state-of-the-art object trackers on three short-term benchmarks (GOT-10K, TrackingNet and VOT2020) and two long-term benchmarks (LaSOT and VOT2020-LT).

{\textbf{GOT-10K.}} GOT-10K~\cite{GOT10K} is a large-scale benchmark covering a wide range of common challenges in object tracking. GOT-10K requires trackers to only use the training set of GOT-10k for model learning. We follow this policy and retrain our models only with the GOT-10K train set. As reported in Tab.~\ref{tab-got10k}, with the same ResNet-50 backbone, STARK-S50 and STARK-ST50 outperform PrDiMP50~\cite{PrDiMP} by 3.8\% and 4.6\% AO scores, respectively. Furthermore, STARK-ST101 obtains a new state-of-the-art AO score of 68.8\%, surpassing Siam R-CNN~\cite{SiamRCNN} by 3.9\% with the same ResNet-101 backbone.

\begin{figure}[!t]
\centering
\begin{minipage}[b]{1\linewidth}
\centering
  \includegraphics[width=0.9\linewidth]{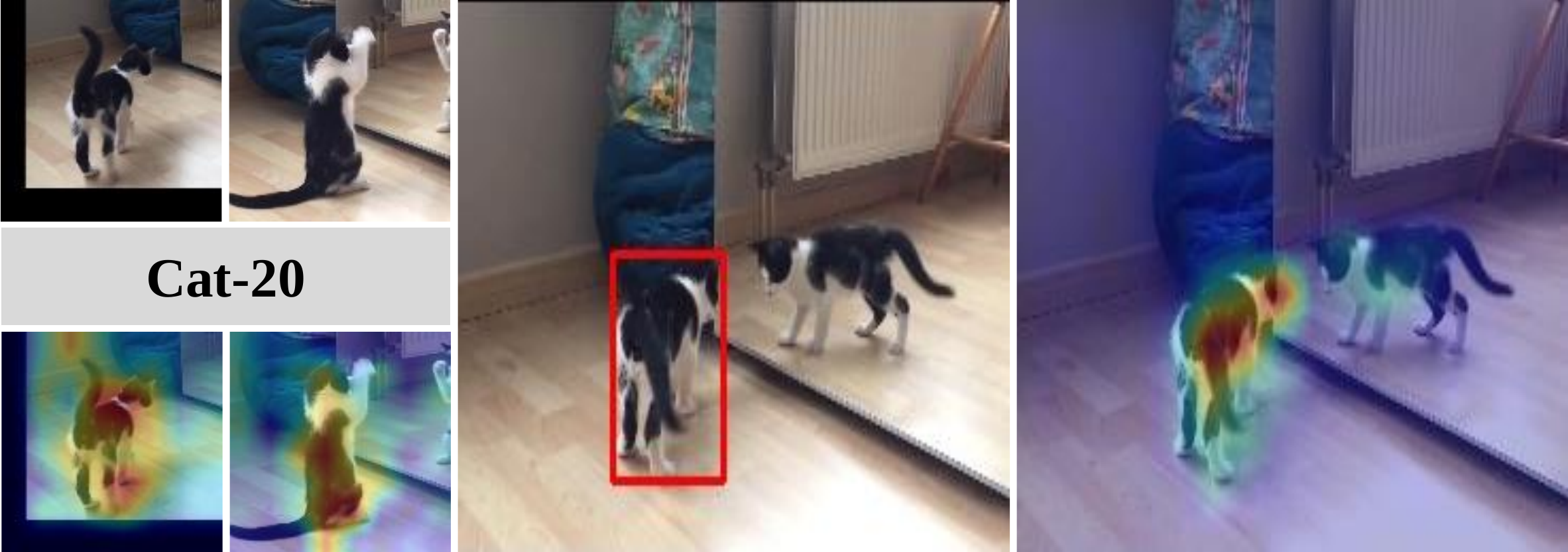} \\
  \includegraphics[width=0.9\linewidth]{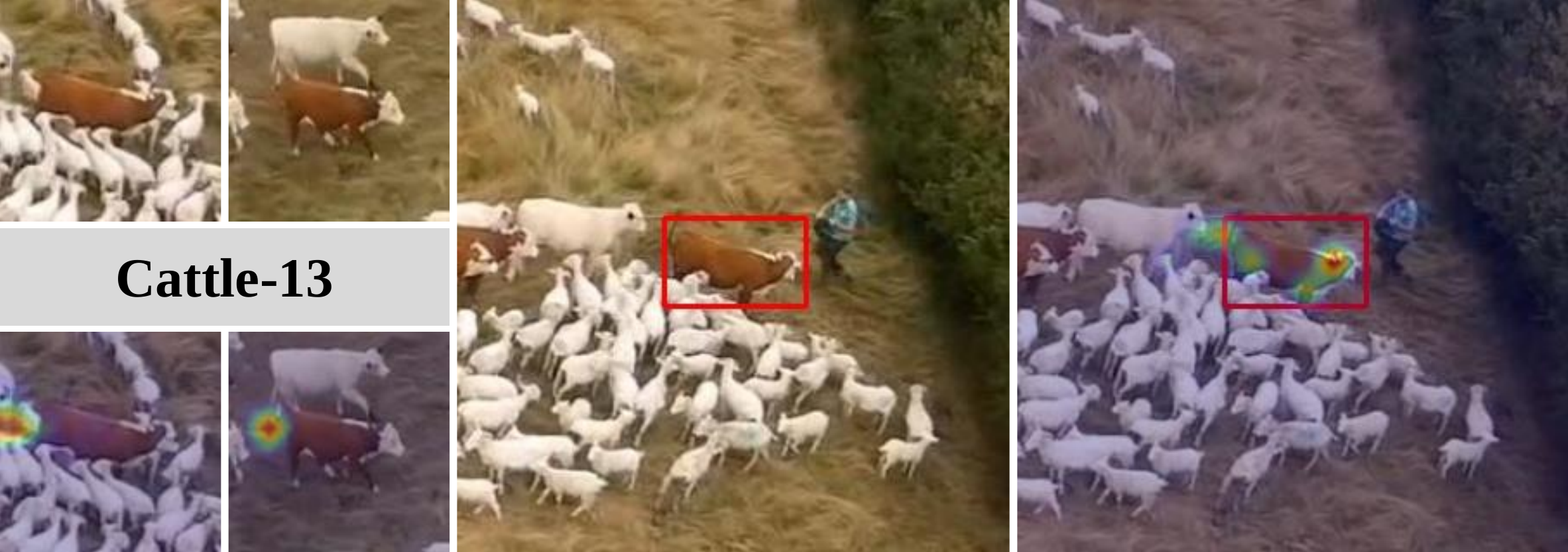}
  \end{minipage}
\caption{Visualization of the encoder attention and the decoder attention.}
\vspace{-5mm}
\label{fig-att}
\end{figure}

{\textbf{TrackingNet.}} TrackingNet~\cite{trackingnet} is a large-scale short-term tracking benchmark containing 511 video sequences in the test set. Tab.~\ref{tab-trackingnet} presents that STARK-S50 and STARK-ST50 surpass PrDiMP50~\cite{PrDiMP} by 4.5\% and 5.5\% in AUC respectively. With a more powerful ResNet-101 backbone, STARK-ST101 achieves the best AUC of 82.0\%, outperforming Siam R-CNN by 0.8\%.

{\textbf{VOT2020.}} Different from previous reset-based evaluations \cite{VOT2019}, VOT2020~\cite{vot2020} proposes a new anchor-based evaluation protocol and uses binary segmentation masks as the groundtruth. The final metric for ranking is the Expected Average Overlap (EAO). Tab.~\ref{tab-vot} shows that STARK-S50 achieves a competitive result, which is better than DiMP~\cite{DiMP} and UPDT~\cite{UPDT}. After introducing temporal information, STARK-ST50 obtains an EAO of 0.308, being superior to previous bounding-box trackers. Inspired by AlphaRef~\cite{vot2020}, the winner of VOT2020 real-time challenge, we equip STARK with a refinement module proposed by AlphaRef to generate segmentation masks. The new tracker ``STARK-ST50+AR" surpasses previous SOTA trackers, like AlphaRef and OceanPlus \cite{Ocean}, getting an EAO of 0.505.

{\textbf{LaSOT.}} LaSOT~\cite{LaSOT} is a large-scale long-term tracking benchmark, which contains 280 videos with average length of 2448 frames in the test set. STARK-S50 and STARK-ST50 achieve a gain of 6.0\% and 6.6\% over PrDiMP~\cite{PrDiMP} respectively, using the same ResNet-50 backbone. Furthermore, STARK-ST101 obtains a success of 67.1\%, which is 2.3\% higher than Siam R-CNN~\cite{SiamRCNN}, as shown in Fig.~\ref{fig-lasot}.

{\textbf{VOT2020-LT.}} VOT2020-LT consists of 50 long videos, in which target objects disappear and reappear frequently. Besides, trackers are required to report the confidence score of the target being present. Precision (Pr) and Recall (Re) are computed under a series of confidence thresholds. F-score, defined as $F=\frac{2PrRe}{Pr+Re}$, is used to rank different trackers. Since STARK-S does not predict this score, we do not report its result on VOT2020-LT. Tab.~\ref{tab-vot20lt} demonstrates that STARK-ST50 and STARK-ST101 outperform all previous methods with an F-score of 70.2\% and 70.1\%, respectively. It is also worth noting that the framework of STARK is much simpler than that of LTMU\_B, the winner of VOT2020-LT Challenge. To be specific, LTMU\_B takes the combination of ATOM~\cite{ATOM} and SiamMask~\cite{SiamMask} as the short-term tracker, MDNet~\cite{MDNet} as the verifier, and GlobalTrack~\cite{GlobalTrack} as the global detector. Whereas there is only one network in STARK and the result is obtained in one forward pass without post-processing.


{\textbf{Speed, FLOPs and Params.}} As demonstrated in Tab.~\ref{tab-speed}, STARK-S50 can run in real-time at more than 40 \emph{fps}. Besides, the FLOPs and Params of STARK-S50 are $4\times$ and $2\times$ less than those of SiamRPN++. Although STARK-ST50 takes a dynamic template as the extra input and introduces an additional score head, the increases of FLOPs and Params is a little, even negligible. This shows that our method can exploit temporal information in a nearly cost-free fashion. When ResNet-101 is used as the backbone, both FLOPs and Params increase significantly but STARK-ST101 can still run at real-time speed, which is 6x faster than Siam R-CNN (5 \emph{fps}), as shown in Fig.~\ref{fig-page1}.

\vspace{-2mm}
\begin{table}[!t]
\small
    \centering
    \caption{\small Comparison about the speed, FLOPs and Params.}
    \vspace{-1mm}
    \begin{tabular}{c|ccc}
        \hline
        \text{Trackers}&Speed(\emph{fps})&FLOPs(G)&Params(M)\\
        \hline
        STARK-S50&42.2&10.5&23.3\\
        STARK-ST50&41.8&10.9&23.5\\
        STARK-ST101&31.7&18.5&42.4\\
        \hline
        SiamRPN++&35.0&48.9&54.0\\
        \hline
    \end{tabular}
    \label{tab-speed}
\vspace{-1mm}
\end{table}

\subsection{Component-wise Analysis}\label{sec-ablation}
In this section, we choose STARK-ST50 as the base model and evaluate the effects of different components in it on LaSOT~\cite{LaSOT}. For simplicity, encoder, decoder, positional encoding, corner prediction, and score head are abbreviated as enc, dec, pos, corner, and score respectively. As shown in Tab.~\ref{tab-component} \text{\#}1, when the encoder is removed, the success drops significantly by 5.3\%. This illustrates that the deep interaction among features from the template and the search region plays a key role. The performance drops by 1.9\% when the decoder is removed as shown in \text{\#}2. This drop is less than that of removing the encoder, showing that the importance of the decoder is less than the encoder. When the positional encoding is removed, the performance only drops by 0.2\% as shown in \text{\#}3. Thus we conclude that the positional encoding is not a key component in our method. We also try to replace the corner head with a three-layer perceptron as in DETR~\cite{DETR}. \text{\#}4 shows that the performance of STARK with an MLP as the box head is 2.7\% lower than that of the proposed corner head. It demonstrates that the boxes predicted by the corner head are more accurate. As shown in \text{\#}5, when removing the score head, the performance drops to 64.5\%, which is lower than that of STARK-S50 without using temporal information. This demonstrates that improper uses of temporal information may hurt the performance and it is important to filter out unreliable templates.  
\begin{table}[!t]
\small
    \centering
    \caption{\small Ablation for important components. Blank denotes the component is used by default, while \xmark\ represents the component is removed. Performance is evaluated on LaSOT.}
    \vspace{-1mm}
    \small
    \begin{tabular}{c|ccccc|c}
        \hline
        \text{\#}&Enc&Dec&Pos&Corner&Score&Success\\
        \hline
        1&\xmark&&&&&61.1\\
        2&&\xmark&&&&64.5\\
        3&&&\xmark&&&66.2\\
        4&&&&\xmark&&63.7\\
        5&&&&&\xmark&64.5\\
        6&&&&&&66.4\\
        \hline
    \end{tabular}
    \label{tab-component}
\end{table}

\begin{table}[!t]
\small
    \centering
    \caption{\small Comparison between STARK and other candidate frameworks. Performance is evaluated on LaSOT.}
    \vspace{-1mm}
    \begin{tabular}{cccccc}
        \hline
        &\tabincell{c}{Template\\query}&\tabincell{c}{Hungarian\\\quad}&\tabincell{c}{Update\\query}&\tabincell{c}{Loc-Cls\\Joint}&\tabincell{c}{Ours\\\quad} \\
        \hline
        Success&61.2&63.7&64.8&62.5&66.4\\
        \hline
    \end{tabular}
    \label{tab-otherframework}
\vspace{-3mm}
\end{table}


  



\vspace{-2mm}
\subsection{Comparison with Other Frameworks}\label{sec-other-framework}
\vspace{-1mm}
In this section, we choose the STARK-ST50 as our base model and compare it with other possible candidate frameworks. These frameworks include generating queries from the template, using the Hungarian algorithm, updating queries as in TrackFormer~\cite{Trackformer}, and jointly learning localization and classification. Due to the space limitation, the figures of the detailed architectures are presented in the \textit{supplementary material}.

\textbf{Template images serve as the queries.} Queries and templates have similar functions in transformer tracking. For example, both of them are expected to encode information about the target objects. From this view, a natural idea is to use template images to serve as the queries of the decoder. Specifically, first, the template and the search region features are separately fed to a weight-shared encoder then the queries generated from the template features are used to interact with the search region features in the decoder. As shown in Tab.~\ref{tab-otherframework}, the performance of this framework is 61.2\%, which is 5.2\% lower than that of our design. We conjecture that the underlying reason is that compared with our method, this design lacks the information flow from the template to the search region, thus weakening the discriminative power of the search region features.

\textbf{Using the Hungarian algorithm \cite{DETR}.} We also try to use $K$ queries, predicting $K$ boxes with confidence scores. $K$ is equal to 10 in the experiments. The groundtruth is dynamically matched with these queries during the training using the Hungarian algorithm. We observe that this training strategy leads to the ``Matthew effect". To be specific, some queries predict slightly more accurate boxes than the others at the beginning of training. Then they are chosen by the Hungarian algorithm to match with the groundtruth, which further enlarges the gaps between the chosen queries and the unselected ones. Finally, there are only one or two queries having the ability to predict high-quality boxes. If they are not selected during inference, the predicted box may become unreliable. As shown in Tab.~\ref{tab-otherframework}, this strategy performs inferior to our method and the gap is 2.7\%. 

\textbf{Updating the query embedding.} Different from STARK exploiting temporal information by introducing an extra dynamic template, TrackFormer~\cite{Trackformer} encodes temporal information by updating the query embedding. Following this idea, we extend the STARK-S50 to a new temporal tracker by updating the target query. Tab.~\ref{tab-otherframework} shows that this design achieves a success of 64.8\%, which is 1.6\% lower than that of STARK-ST50. The underlying reason might be that the extra information brought by an updatable query embedding is much less than that by an extra template. 

\textbf{Jointly learning of localization and classification.} As mentioned in Sec~\ref{sec-temporal}, localization is regarded as the primary task and is trained in the first stage. While classification is trained in the second stage as the secondary task. We also make an experiment to jointly learn localization and classification in one stage. As shown in Tab.~\ref{tab-otherframework}, this strategy leads to a sub-optimal result, which is 3.9\% lower than that of STARK. Two potential reasons are: (1) Optimization of the score head interferes with the training of the box head, leading to inaccurate box predictions. (2) Training of these two tasks requires different data. To be specific, the localization task expects that all search regions contain tracked targets to provide strong supervision. By contrast, the classification task expects a balanced distribution, half of search regions containing the targets, while the remaining half not.

\vspace{-1mm}
\subsection{Visualization}\label{sec-att}

\textbf{Encoder Attention.} The upper part of the Fig.~\ref{fig-att} shows a templates-search triplet from the \textit{Cat-20}, as well as the attention maps from the last encoder layer.
The visualized attention is computed by taking the central pixel of the initial template as the \textit{query} and taking all pixels from the triplet as the \textit{key} and \textit{value}
It can be seen that the attentions concentrate on the tracked target and have roughly separated it from the background. Besides, the features produced by the encoder also have strong discrimination power between the target and distractors.  

\textbf{Decoder Attention.} The lower part of the Fig.~\ref{fig-att} demonstrates a templates-search triplet from the \textit{Cattle-13}, as well as the attention maps from the last decoder layer. It can be seen that decoder attention on templates and search regions are different. Specifically, attention on the templates mainly focuses on the top-left region of the target, while attention on the search region tends to be put on the boundaries of the target. Besides, the learnt attention is robust to distractors.

\vspace{-1mm}
\section{Conclusion}
This paper proposes a new transformer-based tracking framework, which can capture the long-range dependency in both spatial and temporal dimensions. Besides, the proposed STARK tracker gets rid of hyper-parameters sensitive post-processing, leading to a simple inference pipeline. Extensive experiments show that the STARK trackers perform much better than previous methods on five short-term and long-term benchmarks, while running in real-time. 
We expect this work can attract more attention on transformer architecture for visual tracking.

{\small
\bibliographystyle{ieee_fullname}
\bibliography{egbib}
}

\end{document}